\documentclass[10pt,twocolumn,letterpaper]{article}

\usepackage{iccv}
\usepackage{times}
\usepackage{epsfig}
\usepackage{graphicx}
\usepackage{amsmath}
\usepackage{amssymb}
\usepackage{multirow}
\usepackage{subcaption}
\usepackage{soul}
\usepackage{color}
\usepackage{colortbl}
\usepackage{amsthm}
\usepackage[]{algorithm2e}

\usepackage[font=small]{caption}
\definecolor{myRed}{rgb}{0.0, 0.0, 0.0}
\definecolor{myblue}{rgb}{0, .0, 1.0}
\definecolor{ping}{RGB}{100,200,200}

\usepackage[breaklinks=true,bookmarks=false]{hyperref}

\iccvfinalcopy 

\def\iccvPaperID{3242} 

\ificcvfinal\fi
\begin{document}

\title{Batch DropBlock Network for Person Re-identification and Beyond}
\vspace{-2mm}
\author{Zuozhuo Dai$^{1}$\hspace{1cm}Mingqiang Chen$^{1}$\hspace{1cm}Xiaodong Gu$^{1}$\hspace{1cm}Siyu Zhu$^{1}$\hspace{1cm}Ping Tan$^{2}$\\
${}^{1}$Alibaba A.I. Labs\hspace{1.5cm}${}^{2}$Simon Fraser University
}
\maketitle
\footnotetext[1]{Siyu Zhu is the corresponding author.}
\pagestyle{plain}

\begin{abstract}

Since the person re-identification task often suffers from the problem of pose changes and occlusions, some attentive local features are often suppressed when training CNNs. 
In this paper, we propose the Batch DropBlock (BDB) Network which is a two branch network composed of a conventional ResNet-50 as the global branch and a feature dropping branch.
The global branch encodes the global salient representations.
Meanwhile, the feature dropping branch consists of an attentive feature learning module called Batch DropBlock,
which randomly drops the same region of all input feature maps in a batch to reinforce the attentive feature learning of local regions.
The network then concatenates features from both branches and provides a more comprehensive and spatially distributed feature representation.
Albeit simple, our method achieves state-of-the-art on person re-identification and it is also applicable to general metric learning tasks. For instance, we achieve 76.4\% Rank-1 accuracy on the CUHK03-Detect dataset and 83.0\% Recall-1 score on the Stanford Online Products dataset, outperforming the existing works by a large margin (more than 6\%).

\end{abstract}

\section{Introduction}
Person re-identification (re-ID) amounts to identify the same person from multiple detected pedestrian images, typically seen from different cameras without view overlap. It has important applications in surveillance and presents a significant challenge in computer vision. Most of recent works focus on learning suitable feature representation that is robust to pose, illumination, and view angle changes to facilitate person re-ID using convolution neural networks. 
Because the body parts such as faces, hands and feet are unstable as the view angle changes, the CNN tends to focus on the main body part and the other descriminative body parts are consequently suppressed. To solve this problem, many pose-based works \cite{kumar2017pose,su2017pose,suh2018part,zheng2017pose,zhao2017deeply} seek to localize different body parts and align their associated features, and other part-based works \cite{cheng2016person,li2018harmonious,liu2017end,liu2017hydraplus,sun2017beyond,varior2016lstm,xiao2017joint} use coarse partitions or attention selection network to improve feature learning. 
However, such pose-based networks usually require additional body pose or segment information.
Moreover, these networks are designed using specific partition mechanisms, such as a horizontal partition, which is fit for person re-ID but hard to be generalized to other metric learning tasks. The problems above motivate us to propose a simple and generalized network for person re-ID and other metric learning tasks. 

\begin{figure}[t]
    \centering
    \includegraphics[width=1.0\linewidth]{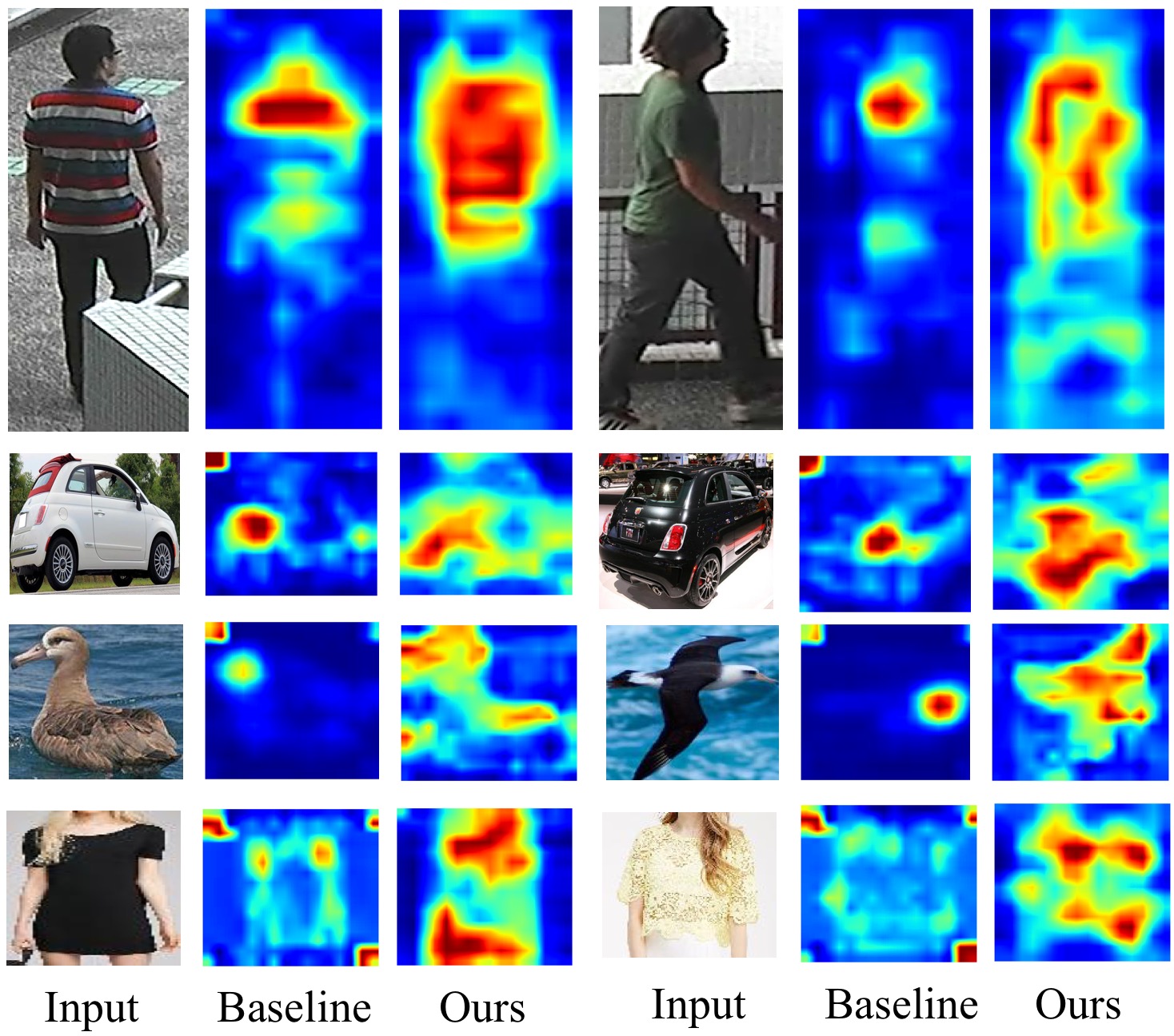}
    \caption{The class activation map on Baseline and BDB Network. Compared with the Baseline, the two-branch structure in BDB Network learns more comprehensive and spatially distributed features \textcolor{myRed}{consisting} of both global and attentive local representations.}
    \label{fig:attention}
\end{figure}

In this paper, we propose the Batch DropBlock Network (BDB Network) for the roughly aligned metric learning tasks. 
The Batch DropBlock Network is a two-branch network \textcolor{myRed}{consisting} of a conventional global branch and a feature dropping branch where the Batch DropBlock, an attentive feature learning module, is applied. The global branch encodes the global feature representations and the feature dropping branch learns local detailed features.
Specifically, Batch DropBlock randomly drops the same region of all the feature maps, namely the same semantic body parts, in a batch during training and reinforces the attentive feature learning of the \textcolor{myRed}{remaining} parts.
Concatenating the features of both branches brings a more comprehensive saliency representation rather than few discriminative features.
In Figure~\ref{fig:attention}, we use class activation map~\cite{zhou2016learning} to visualize the feature attention. We can see that the attention of baseline mainly focuses on the main body part while the BDB network learns more uniformly distributed representations.

Our Batch DropBlock is different from the general DropBlock~\cite{ghiasi2018dropblock} in two aspects.
First, Batch DropBlock is an attentive feature learning module for metric learning tasks while DropBlock is a regularization method for classification tasks.
Second, Batch DropBlock drops the same block for a batch of images during a single iteration, while DropBlock~\cite{ghiasi2018dropblock} erases randomly across different images. 
Here, `Batch' means the group of images participating in a single loss calculation during training, for example, a pair for pairwise loss, a triplet for triplet loss and a quadruplet for quadruplet loss.
If we erase features randomly as~\cite{ghiasi2018dropblock}, for example, one image keeps head features and another image keeps feet features, the network can hardly find the semantic correspondence, not to mention reinforcing the learning of local attentive representations.

In the experimental section, the ResNet-50~\cite{he2016resnet} based Batch DropBlock Network with hard triplet loss~\cite{hermans2017defense} achieves 72.8\% Rank-1 accuracy on CUHK03-Detect dataset, which is 6.0\% higher than the state-of-the-art work~\cite{wang2018mgn}.
Batch DropBlock can also be adopted in different metric learning schemes, including triplet loss~\cite{schroff2015facenet,hermans2017defense}, lifted structure loss~\cite{oh2016deep}, weighted sampling based margin loss~\cite{wu2017sampling}, and histogram loss~\cite{ustinova2016histogram}. We test it with the image retrieval tasks on the CUB200-2011~\cite{wah2011caltech}, CARS196~\cite{krause20133d}, In Shop Clothes Retrieval dataset~\cite{liu2016deepfashion} and Stanford online products dataset~\cite{song2017deep}. The BDB Network can consistently improve the Rank-1 accuracy of various schemes.

\setlength{\textfloatsep}{1pt}

\begin{figure}
\includegraphics[width=1.0\linewidth]{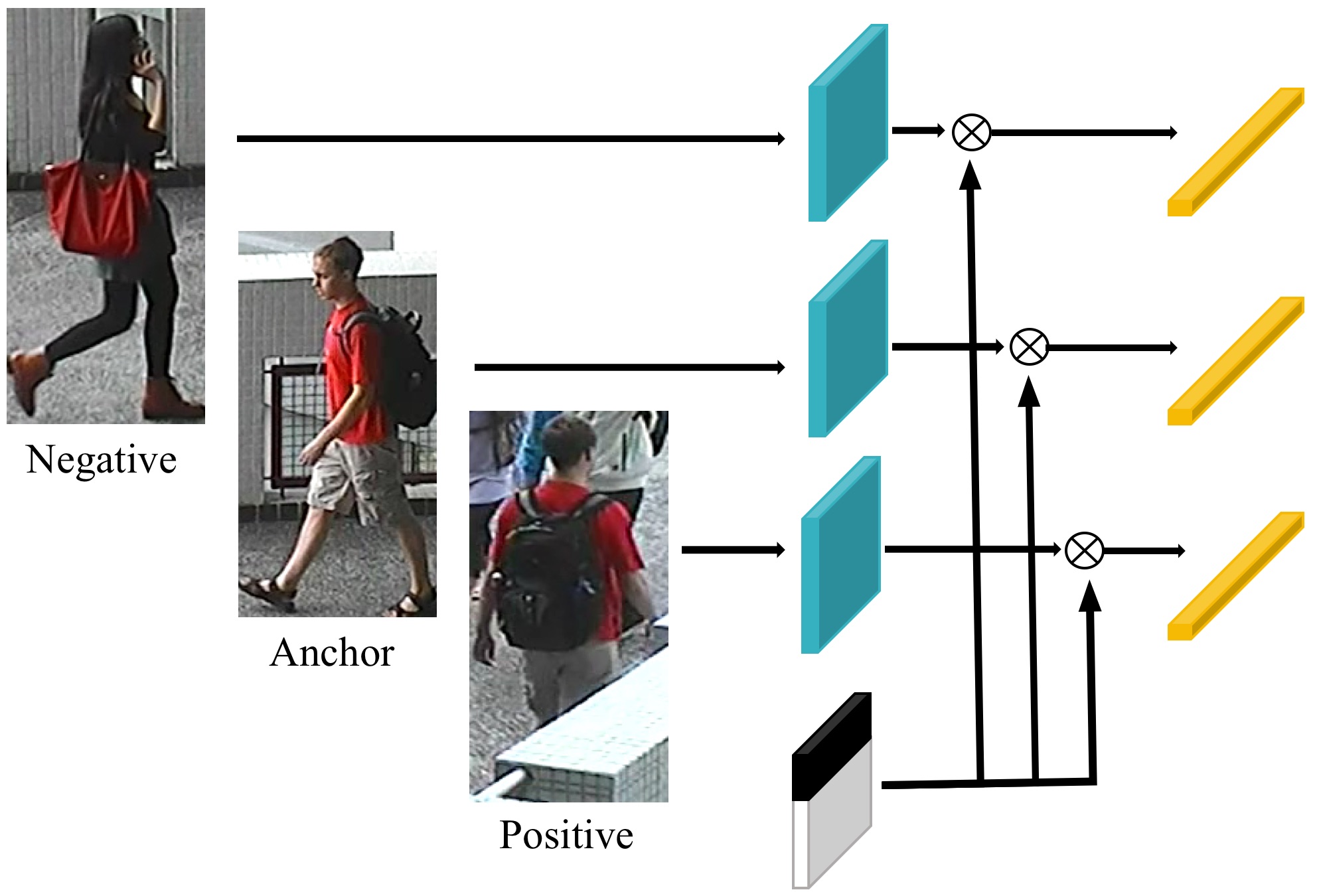}
\caption{The Batch DropBlock Layer demonstrated on the triplet loss function~\cite{schroff2015facenet}.}
\label{fig:triplet}
\vspace{4mm}
\end{figure}

\begin{figure*}[h!]
\begin{center}
\includegraphics[width=0.95\linewidth]{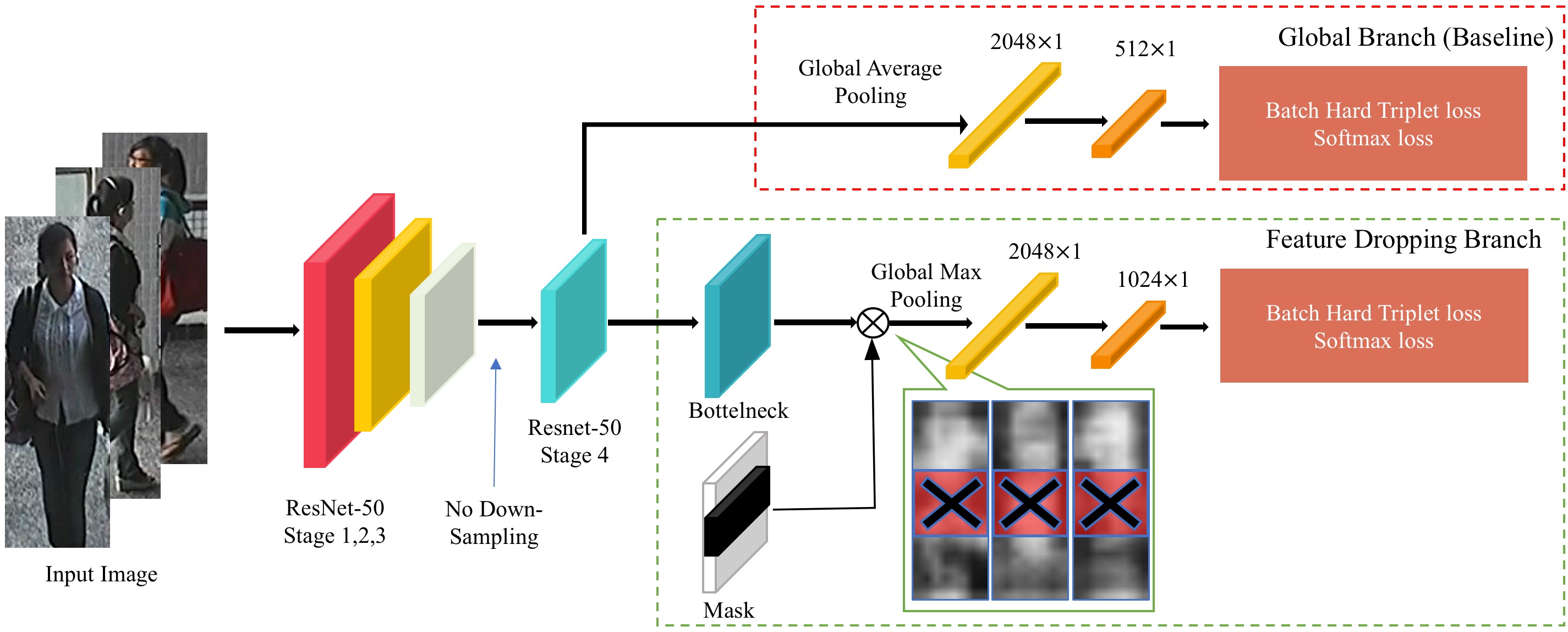}
\end{center}
\vspace{-4mm}
\caption{The structure of our Batch DropBlock (BDB) Network with the batch hard triplet loss~\cite{hermans2017defense} demonstrated on the person re-ID problem.
The global branch is appended after ResNet-50 Stage 4 and the feature dropping branch introduces a mask to crop out a large block in the bottleneck feature map. \textcolor{myRed}{During training, there are two loss functions for both global branch and feature dropping branch. During testing, the features from both branches} are concatenated as the final descriptor of a pedestrian image.}
\label{fig:bfe}
\vspace{-3mm}
\end{figure*}

\section{Related work}
Person re-ID is a challenging task in computer vision due to the large variation of poses, background, illumination, and camera conditions. 
Historically, people used hand-craft features for person re-identification \cite{bazzani2010multishot,das2014consistent,li2013local,liao2015person,ma2013domain,jurie2019pcca,pedagadi2013local,perina2010symmetry,yang2014salient,zheng2013reidentification}.
Recently, deep learning based methods dominate the Person re-ID benchmarks~\cite{chen2018group,shen2018group,sun2017svdnet,zhao2017deeply,zheng2016binary,zheng2017discriminatively}.

The formulation of person re-ID has gradually evolved from a classification problem to a metric learning problem, which aims to find embedding features for input images in order to measure their semantic similarity.
The work~\cite{zheng2016person} compares both strategies on the Market-1501 dataset.
Current works in metric learning generally focus on the design of loss functions, such as contrastive loss~\cite{varior2016contrastive}, triplet loss~\cite{cheng2016person,liu2017end}, lifted structure loss~\cite{oh2016deep}, quadruplet loss~\cite{chen2017beyond}, histogram loss~\cite{ustinova2016histogram}, etc. 
In addition to loss functions, the hard sample mining methods, such as distance weighted sampling~\cite{wu2017sampling}, hard triplet mining~\cite{hermans2017defense} and margin sample mining~\cite{xiao2017margin} are also critical to the final retrieval precision. 
Another work~\cite{zhang2017alignedreid} also studies the application of mutual learning in metric learning tasks.
In this paper, the proposed two-branch BDB Network is effective in many metric learning formulations with different loss functions. 

The human body is highly structured and distinguishing corresponding body parts can effectively determine the identity. Many recent works~\cite{liu2017end,sun2017beyond,ustinova2015bilinear,varior2016lstm,wang2018mgn,wei2017glad,yao2017deep,zhang2017alignedreid,zhao2017spindle} aggregate salient features from different body parts and global cues for person re-ID. Among them, the part-based methods~\cite{cheng2016person,sun2017beyond,wang2018mgn} achieve the state-of-the-art performance, which split an input feature map horizontally into a fixed number of strips and aggregate features from those strips. However, aggregating the feature vectors from multiple branches generally results in a complicated network structure.
In comparison, our method involves only a simple network with two branches, one-third the size of the state-of-the-art MGN method~\cite{wang2018mgn}.

To handle the imperfect bounding box detection and body part misalignment, many works~\cite{li2018harmonious,shen2018group,shen2015structure,si2018dualattention,zheng2015partial} exploit the attention mechanisms to capture and focus on attentive regions.
Saliency weighting~\cite{wang2014salincy,zhao2013unsupervise} is another effective approach to this problem.
Inspired by attention models, Zhao et al.~\cite{zhao2017deeply} propose part-aligned representations for person re-ID.
Following the similar ideology, the works~\cite{kim2018attention,lan2017align,li2017learning,liu2017hydraplus} have also demonstrated superior performance, which incorporate a regional attention selection sub-network into the person re-ID model.
To learn a feature representation robust to pose changes, the pose guided attention methods~\cite{kumar2017pose,su2017pose,zheng2017pose} fuse different body parts features with the help of pose estimation and human parsing network.
However, such methods based on pose estimation and semantic parsing algorithms are only designed for person re-ID tasks while our approach can be applied to other general metric learning tasks.

To further improve the retrieval precision, re-ranking strategies~\cite{bai2017reid,zhong2017re} and inference with specific person attributes~\cite{schumann2017attribute} are adopted too.
Recent works also introduce synthetic training data~\cite{barbosa2017synthetic}, adversarially occluded samples~\cite{huang2018adversarial} and unlabeled samples generated by GAN~\cite{zheng2017unlabeled} to remarkably augment the variant of input training dataset.
The work in~\cite{geng2016deep} transfers the representations learned from the general classification dataset to address the data sparsity of the person re-ID problems.
Some general data augmentation methods such as Random Erasing~\cite{zhong2017re} and Cutout~\cite{devries2017improved} are also generally used.
Notably, such policies above can be used jointly with our method.

\section{Batch DropBlock (BDB) Network} \label{sec:BFE}
This section describes the structure and components of the proposed Batch DropBlock Network.

\paragraph{Backbone Network.}\vspace{-3mm}
We use the ResNet-50~\cite{he2016resnet} as the backbone network for feature extraction as many of the person re-ID networks. For a fair comparison with the recent works~\cite{sun2017beyond, wang2018mgn}, we also modify the backbone ResNet-50 slightly, in which the down-sampling operation at the beginning of stage 4 is not employed. In this way, we get a larger feature map of size $2048\times24\times8$.

\paragraph{ResNet-50 Baseline.}\vspace{-3mm}
On top of this backbone network, we append a branch denoted as {\bf global branch}. Specifically, after stage 4 of ResNet-50, we employ global average pooling to get a 2048-dimensional feature vector, the dimension of which is further reduced to 512 through a $1\times1$ convolution layer, a batch normalization layer, and a ReLU layer.
We denote the backbone network together with the global branch as {\bf ResNet-50 Baseline} in the following sections. 
The performance of Baseline with or without triplet loss on person re-ID datasets are shown in table~\ref{tab:compare_person}. Our baseline without triplet loss is identical to the baseline used in recent works~\cite{sun2017beyond, wang2018mgn}.

\paragraph{Batch DropBlock Layer.}\vspace{-3mm}
Given the feature tensor $T$ computed by backbone network from a single batch of input images,
the Batch DropBlock Layer randomly drops the same region of tensor $T$. All the units inside the dropping area are zeroed out. We visualize the application of Batch DropBlock Layer in the triplet loss function in Figure~\ref{fig:triplet}, while it can be adopted in other loss functions~\cite{oh2016deep,ustinova2016histogram,wu2017sampling} as well. 
The height and width of the erased region varies from task to task. But in general, the dropping region should be big enough to cover a semantic part of input feature map. Unlike DropBlock~\cite{ghiasi2018dropblock}, there is no need to change the keep probability hyper-parameter during training in Batch DropBlock Layer.



\begin{figure}[t]
\includegraphics[width=1\linewidth]{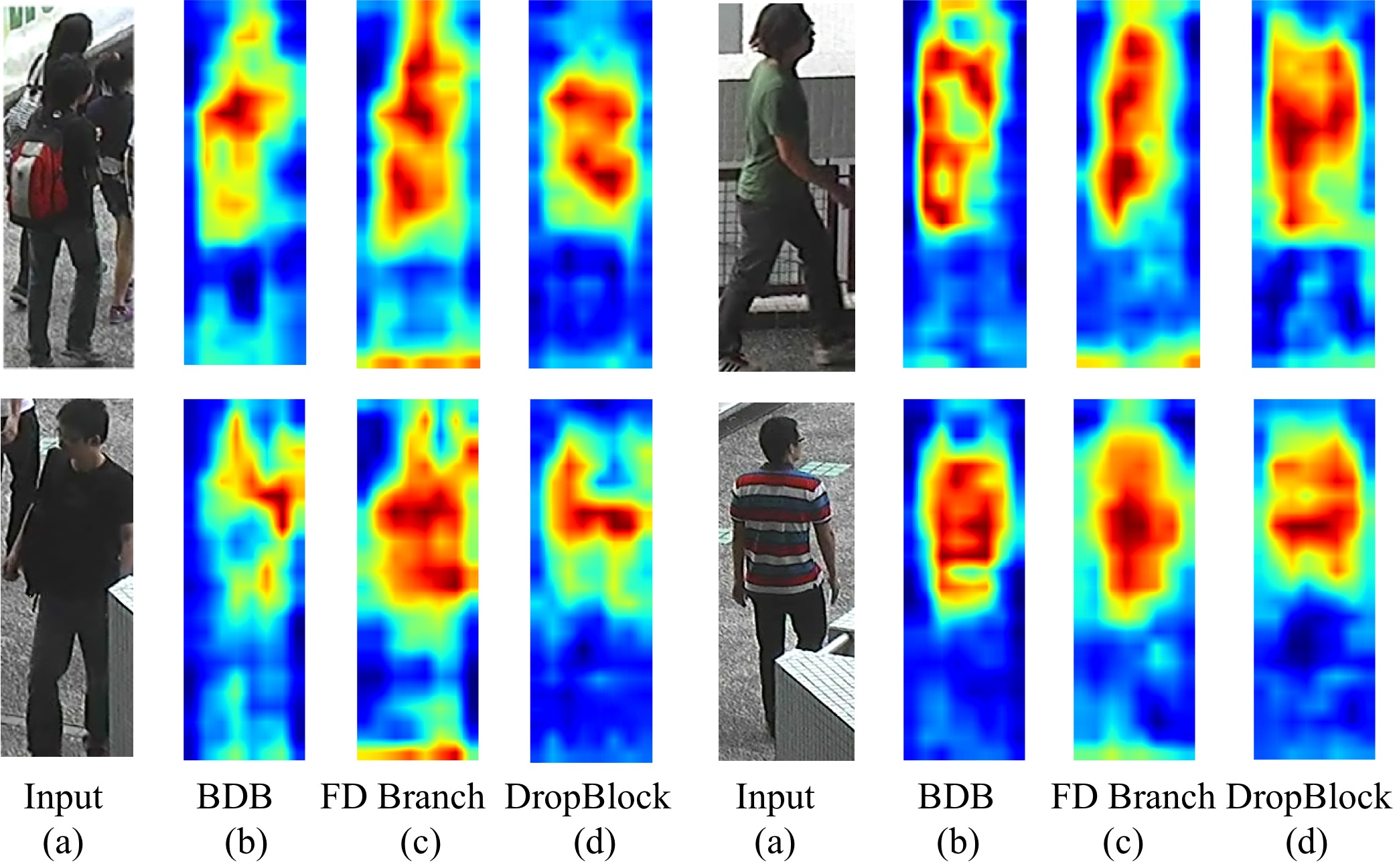}
\vspace{-6mm}
\caption{The class activation map of the BDB Network, the feature dropping branch when training alone, and when DropBlock is used in our network. 'FD Branch' means feature dropping branch.}
\label{fig:attention_bdb}
\vspace{3mm}
\end{figure}

\paragraph{Network Architecture.}\vspace{-3mm}
As illustrated in Figure~\ref{fig:bfe}, our BDB Network consists of a global branch and a feature dropping branch.

The global branch is commonly used for providing global feature representations in multi-branch network architectures~\cite{cheng2016person, sun2017beyond, wang2018mgn}.
It also supervises the training for the feature dropping branch and makes the Batch DropBlock layer applied on a well-learned feature map. 
To demonstrate it, we visualize in Figure~\ref{fig:attention_bdb} the class activation map of the dropping branch trained with and without the global branch. We can see that the features learned by the dropping branch alone are more spatially dispersed with redundant background noise (e.g. at the bottom of Figure~\ref{fig:attention_bdb} (c)). 
As mentioned in~\cite{ghiasi2018dropblock}, dropping a large area randomly on input feature maps may hurt the network learning at the beginning. It therefore uses a scheduled training method which sets the dropping area small initially and gradually increases it to stabilize the training process. In BDB network, we do not need to change the dropping area with the intermediate supervision of the global branch. 
At the beginning stage of training, when the feature dropping branch could not learn well, the global branch helps the training.

The {\bf feature dropping branch} then applies the Batch DropBlock Layer on feature map $T$ and provides the batch erased feature map $T'$. Afterwards, we apply global max pooling to get the 2048-dimensional feature vector. Finally, the dimension of a feature vector is reduced from 2048 to 1024  for both triplet and softmax losses. The purpose of the feature dropping branch is to learn multiple attentive feature regions instead of only focusing on the major discriminative region. Figure~\ref{fig:attention_bdb} also visualizes the class activation map of feature dropping branch with DropBlock or Batch DropBlock. One can see the features learned by DropBlock miss some attentive part features (e.g. legs in Figure~\ref{fig:attention_bdb} (d)) and the salient representations from Batch DropBlock have more accurate and clearer contours.
An intuitive explanation is that, by blocking the same roughly aligned regions, we reinforce the attentive feature learning of the rest parts with semantic correspondences.

The BDB Network uses global average pooling (GAP) on the global branch, the same as the original ResNet-50 network~\cite{he2016resnet}. Notably, we use global max pooling (GMP) in feature dropping branch, because GMP encourages the network to identify comparatively weak salient features after the most descriminative part is dropped.
The strong feature is easy to be selected while the weak feature is hard to be distinguished from other low values. When the strong feature is dropped, GMP could encourage the network to strength the weak features. For GAP, low values except the weak features would still impact the results.

Also noteworthy is the ResNet bottleneck block~\cite{he2016resnet} which applies a stack of convolution layers on feature map $T$.
Without it, the global average pooling layer and the global max pooling layer would be applied simultaneously on $T$, making the network hard to converge.


Then, \textcolor{myRed}{during testing,} features from the global branch and the feature dropping branch are concatenated as the embedding vector of a pedestrian image. Here, the following three points are worth noting.
1) The Batch DropBlock Layer is parameter free and will not increase the network size.
2) The Batch DropBlock Layer can be easily adopted in other metric learning tasks beyond person re-ID.
3) The Batch DropBlock hyper-parameters are tunable without changing the network structure for different tasks.

\paragraph{Loss function.}\vspace{-3mm}
The loss function is the sum of soft margin batch-hard triplet loss~\cite{hermans2017defense} and softmax loss on both the global branch and feature dropping branch. 

\begin{table*}[t]
    \begin{center}
    \resizebox{0.8\linewidth}{!}{
    \begin{tabular}{c|c|c|c|c|c|c|c|c}
     \hline
     & \multicolumn{2}{c|}{CUHK03-Label} & \multicolumn{2}{|c|}{CUHK03-Detect} & \multicolumn{2}{|c|}{DukeMTMC-reID} & \multicolumn{2}{|c}{Market1501} \\
     Method & Rank-1 & mAP & Rank-1 & mAP & Rank-1 & mAP & Rank-1 & mAP \\
     \hline
     IDE~\cite{zheng2016person} & 22.2 & 21.0 & 21.3 & 19.7 & 67.7 & 47.1 & 72.5 & 46.0 \\
     PAN~\cite{zheng2018pedestrian} & 36.9 & 35.0 & 36.3 & 34.0 &71.6 & 51.5 & 82.8 & 63.4 \\
     SVDNet~\cite{sun2017svdnet} & - & - & 41.5 & 37.3 & 76.7 & 56.8 & 82.3 & 62.1 \\
     DPFL~\cite{chen2018person} & 43.0 & 40.5 & 40.7 & 37.0 & 79.2 & 60.0 & 88.9 & 73.1 \\
     HA-CNN~\cite{li2018harmonious} & 44.4 & 41.0 & 41.7 & 38.6 & 80.5 & 63.8 & 91.2 & 75.7 \\
     SVDNet+Era~\cite{zhong2017random} & 49.4 & 45.0 & 48.7 & 37.2 & 79.3 & 62.4 & 87.1 & 71.3 \\
     TriNet+Era~\cite{zhong2017random} & 58.1 & 53.8 & 55.5 & 50.7 & 73.0 & 56.6 & 83.9 & 68.7 \\
     DaRe~\cite{wang2018resource} & 66.1 & 61.6 & 63.3 & 59.0 & 80.2 & 64.5 & 89.0 & 76.0 \\
     GP-reid~\cite{almazan2018re} & - & - & - & - & 85.2 & 72.8 & 92.2 & 81.2 \\
     PCB~\cite{sun2017beyond} & - & - & 61.3& 54.2  & 81.9 & 65.3 & 92.4 & 77.3\\
     PCB + RPP~\cite{sun2017beyond} & - & - & 62.8 & 56.7 & 83.3 & 69.2 & 93.8 & 81.6\\
     MGN~\cite{wang2018mgn} & 68.0 & 67.4 & 66.8 & 66.0 & 88.7 & {\bf 78.4} & {\bf 95.7} & {\bf 86.9} \\
     \hline
     Baseline & 52.6 & 49.9 & 51.1 & 47.9 & 81.0 & 62.8	& 91.6 & 77.1\\
     Baseline+Triplet & 67.4 & 61.5 & 63.6 & 60.0& 83.8 & 68.5 & 93.1 & 80.6\\
     \hline
     BDB & 73.6 & 71.7 & 72.8 & 69.3 & 86.8 & 72.1 & 94.2 & 84.3\\
     BDB+Cut & {\bf 79.4} & {\bf 76.7} & {\bf 76.4} & {\bf 73.5} & {\bf 89.0} & 76.0 & 95.3 & 86.7\\
     \hline
    \end{tabular}
    }
    \end{center}
    \vspace{-5mm}
    \caption{The comparison with the existing person re-ID methods. `Era' means Random Erasing~\cite{zhong2017random}. `Cut' means Cutout~\cite{devries2017improved}.}
    \label{tab:compare_person}
    \vspace{-4mm}
\end{table*}

\section{Experiments}
We verify our BDB Network on the benchmark person re-ID datasets.
The BDB Network with different metric learning loss functions is also tested on the standard image retrieval datasets. 

\subsection{Person re-ID Experiments}
\subsubsection{Datasets and Settings}
We test three generally used person re-ID datasets including Market-1501~\cite{zheng2015scalable}, DukeMTMC-reID~\cite{Ristani2016Performance,zheng2017unlabeled}, and CUHK03~\cite{Li2014DeepReID} datasets. 
We also follow the same strategy used in recent works~\cite{hermans2017defense, sun2017beyond, wang2018mgn} to generate training, query, and gallery data.
Notice that the original CUHK03 dataset is divided into 20 random training/testing splits for cross validation which is commonly used in hand-craft feature based methods. 
The new partition method adopted in our experiments further splits the training and gallery images, and selects challenging query images for evaluation. 
Therefore, CUHK03 dataset becomes the most challenging dataset among the three.

During training, the input images are re-sized to $384\times128$ and then augmented by random horizontal flip and normalization.
In Batch DropBlock layer, we set the erased height ratio $r_{h}$ to 0.3 and erased width ratio $r_{w}$ to 1.0. The same setting is used in all the person re-ID datasets. The testing images are re-sized to $384\times128$ and only augmented with normalization.

For each query image, we rank all the gallery images in decreasing order of their Euclidean distances to the query images and compute the Cumulative Matching Characteristic (CMC) curve. 
We use Rank-1 accuracy and mean average precision (mAP) as the evaluation metrics. 
Results with the same identity and the same camera ID as the the query image are not counted. 
It is worth noting that all the experiments are conducted in a single-query setting without re-ranking\cite{bai2017reid,zhong2017re} for simplicity.
\vspace{-3mm}
\subsubsection{Training}\vspace{-1mm}
Our network is trained using 4 GTX1080 GPUs with a batch size of 128. 
Each identity contains 4 instance images in a batch, so there are 32 identities per batch. 
The backbone ResNet-50 is initialized from the ImageNet~\cite{deng2009imagenet} pre-trained model. 
We use the batch hard soft margin triplet loss~\cite{hermans2017defense} to avoid margin parameters.
We use the Adam optimizer~\cite{kingma2014adame} with the base learning rate initialized to 1e-3 with a linear warm-up~\cite{goyal2017accurate} in first 50 epochs, then decayed to 1e-4 after 200 epochs, and further decayed to 1e-5 after 300 epochs. The whole training procedure has 400 epochs and takes approximately 1.5 hours. 
\vspace{-3mm}
\subsubsection{Comparison with State-of-the-Art}\vspace{-1mm}
The statistical comparison between our BDB Network and the state-of-the-art methods on CUHK03, DukeMTMC-reID and Market-1501 datasets is shown in Table~\ref{tab:compare_person}. It shows that our method achieves state-of-the-art performance on both CUHK03 and DukeMTMC-reID datasets. Remarkably, our method achieves the largest improvement over previous methods on CUHK03-Detect dataset, which is the most challenging dataset. For Market1501 datasets, our model achieves comparative performance to MGN~\cite{wang2018mgn}. However, it is worth to point out that MGN benefits from a much lager and more complex network which generates 8 feature vectors with 8 branches supervised by 11 loss functions. 
The model size (i.e., number of parameters) of MGN is three times of BDB Network.

\begin{figure}[t]
\begin{center}
\includegraphics[width=0.8\linewidth]{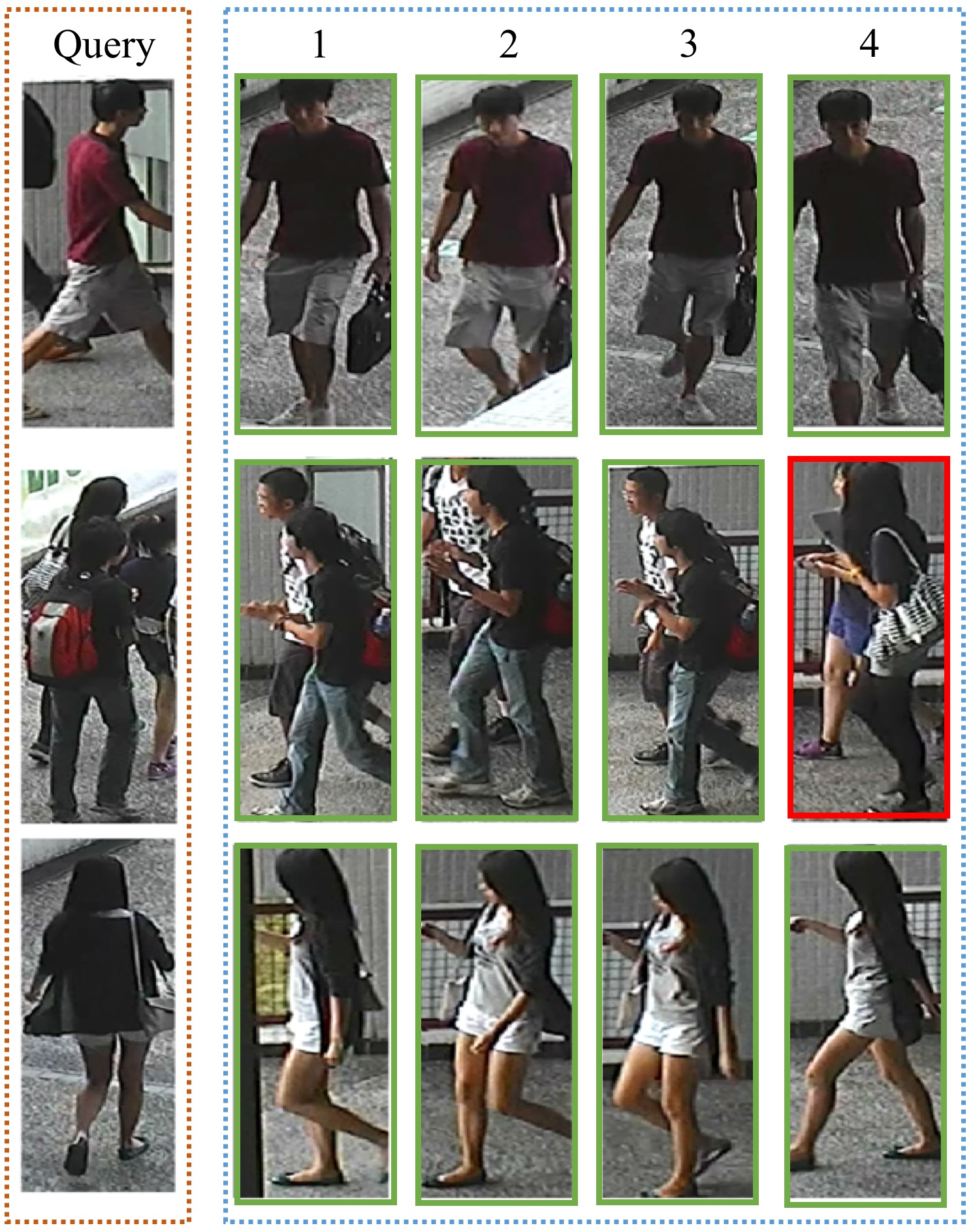}
\end{center}\vspace{-3mm}
\caption{The top-4 ranking list for the query images on CUHK03-Label dataset from the proposed BDB Network. 
The correct results are highlighted by green borders and the incorrect results by red borders.}
\label{fig:cuhk_results}
\vspace{3mm}
\end{figure}

Some sample query results are illustrated in Figure~\ref{fig:cuhk_results}. We can see that, given a back view person image, BDB Network can even retrieve the front view and side view images of the same person.
\vspace{-3mm}
\subsubsection{Ablation Studies}
We perform extensive experiments on Market-1501 and CUHK03 datasets to analyze the effectiveness of each component and the impact of hyper parameters in our method.

\begin{table}[]
    \begin{center}
    \resizebox{0.8\linewidth}{!}{
    \begin{tabular}{c|c|c}
        \hline
         Method & Rank-1 & mAP  \\
         \hline
         Global Branch (Baseline) & 93.1 & 80.6 \\
         Feature Dropping Branch & 93.6 & 83.3 \\
         Both Branches (BDB)    & 94.2  & 84.3 \\
         \hline
         Feature Dropping Branch + Cut & 88.0 & 75.7\\
         BDB + Cut & 95.3 & 86.7\\
         \hline
    \end{tabular}
    }
    \end{center}
    \vspace{-4mm}
    \caption{The effect of global branch and feature dropping branch on Market-1501 dataset. `Cut' means Cutout~\cite{devries2017improved} augmentation.}
    \label{tab:global_branch}
    \vspace{1mm}
\end{table}


\paragraph{Benefit of Global Branch and Feature Dropping Branch.}\vspace{-4mm}
Without the global branch, the BDB Network still performs better than the baseline as illustrated in Table~\ref{tab:global_branch}. Adding the global branch could further improve the performance. The motivation behind the two-branch structure in the BDB Network is that it learns both the most salient appearance clues and fine-grained discriminative features. This suggests that the two branches reinforce each other and are both important to the final performance.

\begin{figure}[t]
    \centering
    \includegraphics[width=1\linewidth]{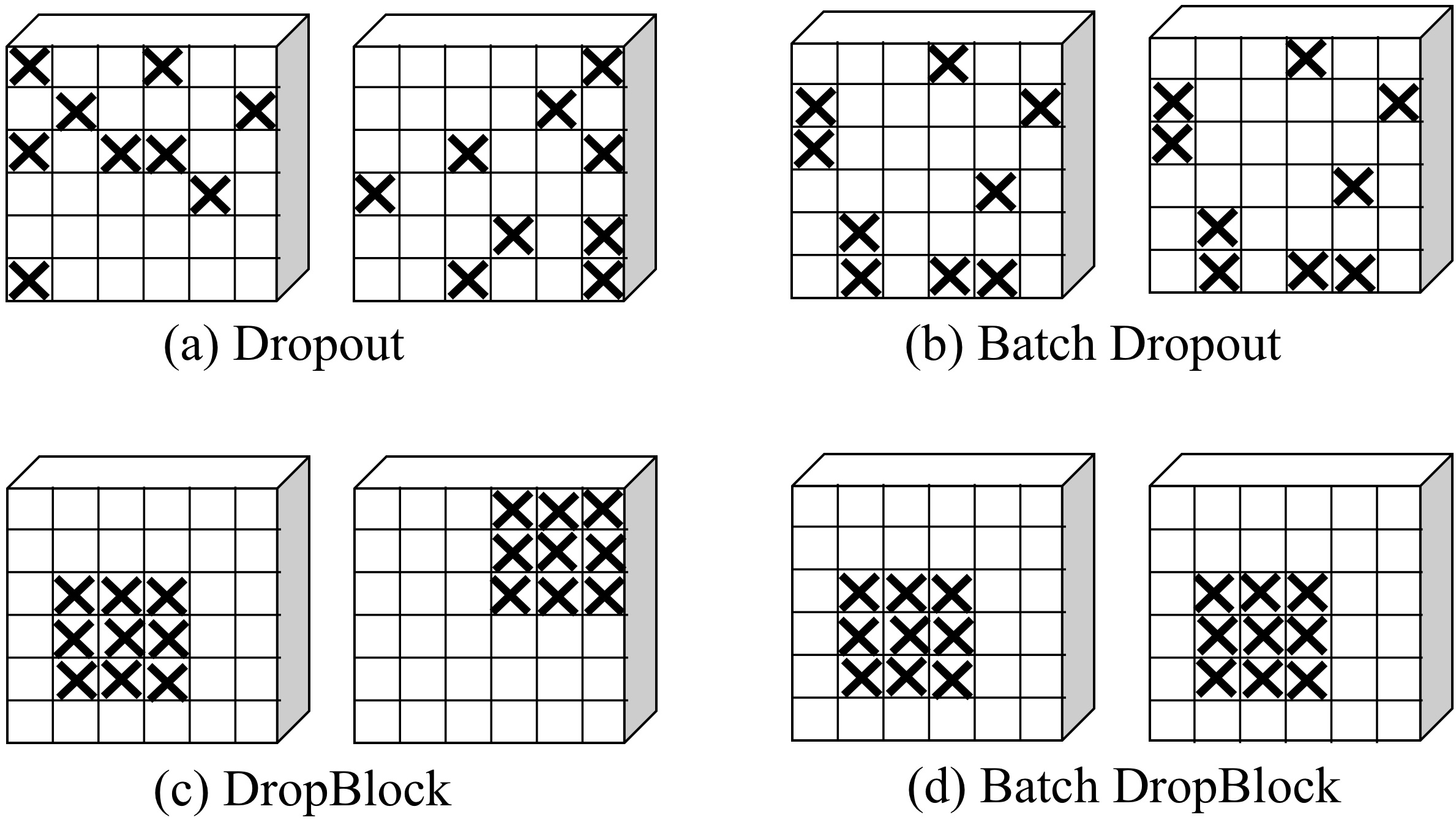}
    \vspace{-3mm}
    \caption{The comparison with Dropout methods on two feature maps within the same batch.}\label{fig:dropout}
    \vspace{3mm}
\end{figure}

\begin{table*}[!htb]
\begin{minipage}[t]{.25\linewidth}
\resizebox{\linewidth}{!}{%
\begin{tabular}{c|c|c}
\hline
Method & Rank-1 & mAP  \\
\hline
SpatialDropout\cite{tompson2015efficient} & 60.5 & 56.8 \\
Dropout~\cite{srivastava2014dropout} & 65.3 & 62.2 \\
Batch Dropout & 65.8 & 62.9 \\
DropBlock~\cite{ghiasi2018dropblock} & 70.6 & 67.7 \\
\hline
Batch DropBlock & 72.8 & 69.3 \\
\hline
\end{tabular}
}
\vspace{0mm}
\caption{The Comparison with other Dropout methods on the CUHK03-Detect dataset.}
\label{table:variants}
\end{minipage}%
\hfill%
\begin{minipage}[t]{.38\linewidth}
\resizebox{\linewidth}{!}{%
\begin{tabular}{c|c|c|c|c}
\hline
& \multicolumn{2}{c|}{CUHK03-Detect} & \multicolumn{2}{|c}{Market1501} \\
Method & Rank-1 & mAP & Rank-1 & mAP \\
     \hline
     Baseline & 51.1 & 47.9 & 91.6 & 77.1\\ 
     Baseline + Triplet & 63.6 & 60.0 & 93.1 & 80.6\\
     Baseline + Dropping & 60.9 & 57.2 & 93.8 & 80.5\\
     \hline
     Baseline + Triplet + &\multirow{2}{*}{72.8} &\multirow{2}{*}{69.3} &\multirow{2}{*}{94.2} &\multirow{2}{*}{84.3}\\
     Dropping (BDB Network) & & & &\\
     \hline
\end{tabular}
}
\caption{Ablation studies of the effective components of BDB network on CUHK03-Detect and Market1501 datasets. `Dropping' means the feature dropping branch.}
\label{table:triplet_loss}
\end{minipage}%
\hfill%
\begin{minipage}[t]{.33\linewidth}
\resizebox{\linewidth}{!}{%
\begin{tabular}{c|c|c|c|c}
\hline
& \multicolumn{2}{c|}{CUHK03-Detect} & \multicolumn{2}{|c}{Market1501} \\
Method & Rank-1 & mAP & Rank-1 & mAP \\
\hline
Baseline & 63.6 & 60.0 & 93.1 & 80.6 \\
Baseline + RE & 70.6 & 65.9 & 93.3 & 81.5\\
Baseline + Cut & 67.7 & 64.2 & 93.5 & 82.0 \\
Baseline + RE + Cut & 70.7 & 65.9 &  93.1 & 82.0 \\
\hline\hline
BDB & 72.8 & 69.3 & 94.2 & 84.3\\
BDB + RE & 75.9 & 72.6 & 94.4 & 85.0 \\
BDB + Cut & {\bf 76.4} & {\bf 73.5} & {\bf 95.3} & {\bf 86.7}\\
\hline
\end{tabular}
}
\vspace{-3mm}
\caption{The comparison with data augmentation methods. `RE' means Random Erasing~\cite{zhong2017random}. `Cut' means Cutout~\cite{devries2017improved}.}
\label{table:augmentation}
\end{minipage} 
\vspace{-3mm}
\end{table*}

\paragraph{Comparison with Dropout and DropBlock.}\vspace{-4mm} Dropout~\cite{srivastava2014dropout} drops values of input tensor randomly and is a widely used regularization technique to prevent overfitting. 
We replace the Batch DropBlock layer with various Dropout methods and compare their performance in Table~\ref{table:variants}. 
SpatialDropout~\cite{tompson2015efficient} randomly zeroes whole channels of the input tensor. The channels to zero-out are randomized on every forward call. 
Here, Batch Dropout means we select random spatial positions and drops all input features in these locations. The difference between Batch DropBlock and Batch Dropout is that Batch DropBlock zeroes a large contiguous area while Batch Dropout zeroes some isolated features. 
DropBlock~\cite{ghiasi2018dropblock} means for a batch of input tensor, every tensor randomly drops a contiguous region. 
The difference between Batch DropBlock and DropBlock is that Batch DropBlock drops the same region for every input tensor within a batch while DropBlock crops out different regions. These Dropout methods are visualized in Figure~\ref{fig:dropout}.
As shown in Table~\ref{table:variants}, Batch DropBlock is more effective than these various Dropout strategies in the person re-ID tasks.

\begin{figure}
\includegraphics[width=1.0\linewidth]{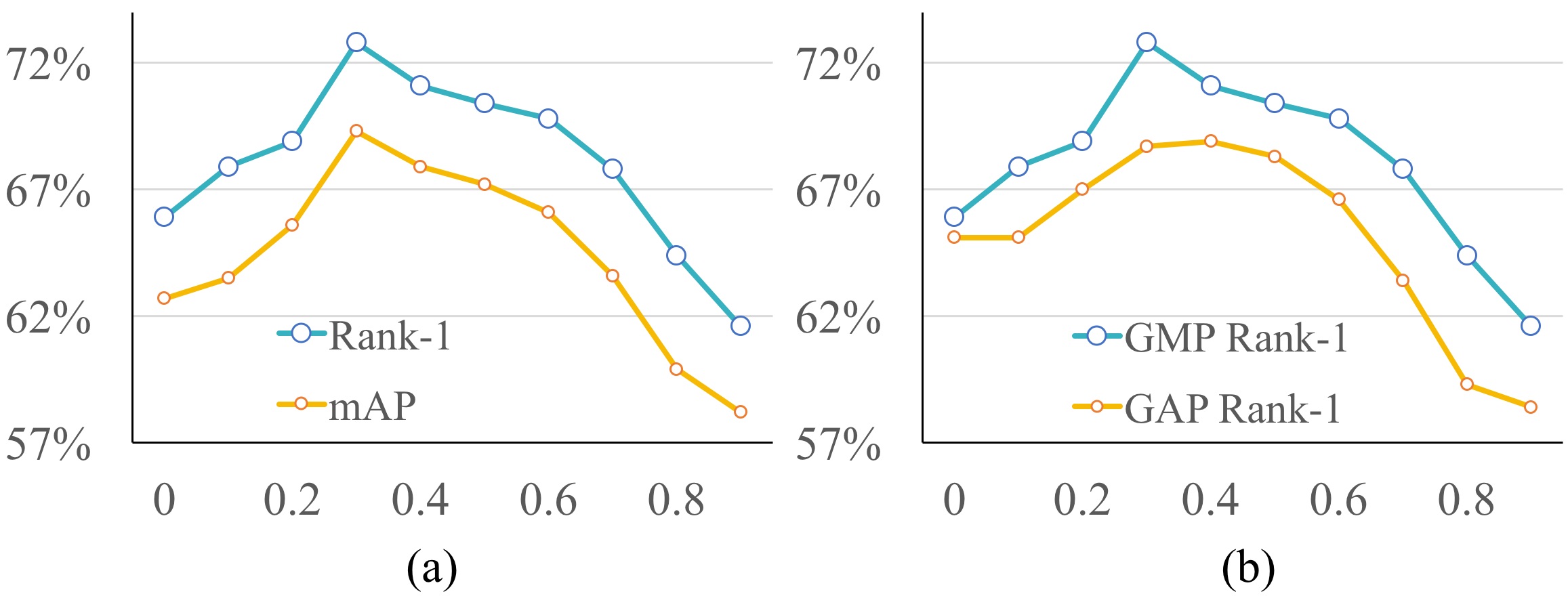}
\vspace{-6mm}
\caption{(a) The effects of erased height ratio on mAP and CMC scores. The erased width ratio is fixed to 1.0. (b) The comparison of  global average pooling and global max pooling on the feature dropping branch under different height ratio settings. The statistics are analyzed on the CUHK03-Detect dataset.}
\label{fig:cuhk_ratio}
\vspace{2mm}
\end{figure}

\paragraph{Global Average Pooling (GAP) vs Global Max Pooling (GMP) in Feature Dropping Branch.}\vspace{-4mm}
As shown in Figure~\ref{fig:cuhk_ratio} (b), the Rank-1 accuracy of the feature dropping branch with GMP is consistently superior to that with GAP.
We therefore demonstrate the importance of Max Pooling \textcolor{myRed}{for} a robust convergence and increased performance on the feature dropping branch.

\paragraph{Benefit of Triplet Loss}\vspace{-4mm}
The BDB Network is trained using both triplet loss and softmax loss. The triplet loss is a vital part of BDB Network since the Batch DropBlock layer has effect only when considering relationship between images. In table~\ref{table:triplet_loss}, `Baseline + Dropping' is the BDB Network without triplet loss. We can see that the triplet loss significantly improves the performance.

\paragraph{Impact of Batch DropBlock Layer Hyper-parameters.}\vspace{-4mm}
Figure~\ref{fig:cuhk_ratio} (a) studies the impact of erased height ratio on the performance of the BDB Network.  
Here, the erased width ratio is fixed to 1.0 in all the person Re-ID experiments.
We can see that the best performance is achieved when height erased ratio is 0.3, which is the setting for BDB Network in person re-ID experiments.


\paragraph{Relationship with Data Augmentation methods.}\vspace{-4mm}
A natural question about BDB Network is could BDB Network still benefit from image erasing data augmentation methods such as Cutout~\cite{devries2017improved} and Random Erasing~\cite{zhong2017random} since they perform similar operations? The answer is yes. Because the BDB Network contains a global branch which sees the complete feature map and it can benefit from Cutout or Random Erasing. 
To verify it, we apply image erasing augmentation on BDB Network with or without the global branch in Table~\ref{tab:global_branch}. We can see Cutout performs bad without the global branch. Table~\ref{table:augmentation} shows BDB Network performs well with data augmentation methods. As can be seen, `BDB + Cut' or `BDB + RE' are significantly better than `Baseline + Cut', `Baseline + RE', or `BDB'.

\begin{table}
\begin{center}
    \resizebox{0.9\linewidth}{!}{
    \begin{tabular}{c|c|c|c|c}
         \hline
         Dataset & CARS & CUB & SOP & Clothes \\
         \hline\hline
         \# images & 16,185 & 11,788 & 120,053 & 52,712 \\
         \# classes & 196 & 200 & 22,634 & 11,735 \\
         \hline\hline
         \# training class & 98 & 100 & 11,318 & 3,997 \\ 
         \# training image & 8,054& 5,864 & 59,551 & 25,882 \\ 
         \hline\hline
         \# testing class & 98 & 100 & 11,316 & 3,985 \\
         \# testing image & 8,131 &5,924 & 60,502 & 26,830 \\
         \hline
    \end{tabular}
    }
    \end{center}
    \vspace{-4mm}
    \caption{The statistics of the image retrieval datastes including CARS196~\cite{krause20133d}, CUB200-2011~\cite{wah2011caltech}, Stanford online products(SOP)~\cite{oh2016deep}, and In-Shop Clothes retrieval dataset~\cite{liu2016deepfashion}. Notice that the test set of In-Shop Clothes retrieval dataset is further split to query dataset with 14,218 images and gallery dataset with 12,612 images.}
    \label{tab:retrieval_dataset}
    \vspace{2mm}
\end{table}

\begin{table*}[!htb]
    \begin{subtable}{.5\linewidth}
      \centering
         \resizebox{0.82\linewidth}{!}{
        \begin{tabular}{c|c|c|c|c}
         \hline
         $K$ & 1 & 2 & 4 & 8 \\
         \hline
         PDDM Triplet~\cite{huang2016local} & 50.9 & 62.1 & 73.2 & 82.5 \\
         PDDM Quadruplet~\cite{huang2016local} & 58.3 & 69.2 & 79.0 & 88.4 \\
         HDC~\cite{yuan2017hard} & 60.7 & 72.4 & 81.9 & 89.2 \\
         Margin~\cite{wu2017sampling} & 63.9 & 75.3 & 84.4 & 90.6  \\
         ABE-8~\cite{kim2018attention} & 70.6 & 79.8 & 86.9 & 92.2 \\
         \hline
         BDB & {\bf 74.1} & {\bf 83.6} & {\bf 89.8} & {\bf 93.6}  \\
         \hline
        \end{tabular}
        }
        \caption{CUB200-2011 (cropped) dataset}
    \end{subtable}
    \hspace{-3mm}
    \begin{subtable}{.5\linewidth}
      \centering
        \resizebox{0.82\linewidth}{!}{
        \begin{tabular}{c|c|c|c|c}
            \hline
            $K$ & 1 & 2 & 4 & 8 \\
            \hline
            PDDM Triplet~\cite{huang2016local} & 46.4 & 58.2 & 70.3 & 80.1 \\
            PDDM Quadruplet~\cite{huang2016local} & 57.4 & 68.6 & 80.1 & 89.4 \\
            HDC~\cite{yuan2017hard} & 83.8 & 89.8 & 93.6 & 96.2 \\
            Margin~\cite{wu2017sampling} & 86.9 & 92.7 & 95.6 & 97.6 \\
            ABE-8~\cite{kim2018attention} & 93.0 & 95.9 & 97.5 & 98.5 \\
            \hline
            BDB & {\bf 94.3} & {\bf 96.8} & {\bf 98.3} & {\bf 98.9} \\
            \hline
       \end{tabular}
       }
        \caption{CARS196 (cropped) dataset}
    \end{subtable} 

    \begin{subtable}{.5\linewidth}
      \centering
        \vspace{3mm}
        \resizebox{0.84\linewidth}{!}{
        \begin{tabular}{c|c|c|c|c|c}
            \hline
            $K$ & 1 & 10 & 20 & 30 & 40   \\
            \hline
            FasionNet~\cite{liu2016deepfashion} & 53.0 & 73.0 & 76.0 & 77.0 & 79.0 \\
            HDC~\cite{yuan2017hard} & 62.1 & 84.9 & 89.0 & 91.2 & 92.3  \\
            DREML~\cite{xuan2018deep} & 78.4 & 93.7 & 95.8 & 96.7 & -  \\
            HTL~\cite{ge2018deep} & 80.9 & 94.3 & 95.8 & 97.2 & 97.4 \\
            A-BIER~\cite{opitz2018deep} & 83.1 & 95.1 & 96.9 & 97.5 & 97.8  \\
            ABE-8~\cite{kim2018attention} & 87.3 & {\bf 96.7} & {\bf 97.9} & 98.2 & 98.5  \\
            \hline
            BDB & {\bf 89.1 } & 96.3 &  97.6  & {\bf 98.5 }& {\bf 99.1 } \\
            \hline
       \end{tabular}
       }
        \caption{In-Shop Clothes Retrieval dataset}
    \end{subtable}
    \hspace{-5mm}
    \begin{subtable}{.5\linewidth}
    \vspace{3mm}
      \centering
        \resizebox{0.77\linewidth}{!}{
        \begin{tabular}{c|c|c|c|c}
            \hline
            $K$ & 1 & 10 & 100 & 1000   \\
            \hline
            LiftedStruct~\cite{oh2016deep} & 62.1 & 79.8 & 91.3 & 97.4   \\
            N-Pairs~\cite{sohn2016improved} & 67.7 & 83.8 & 93.0 & 97.8  \\
            Margin~\cite{wu2017sampling} & 72.7 & 86.2 & 93.8 & 98.0  \\
            HDC~\cite{yuan2017hard} & 69.5 & 84.4 & 92.8 & 97.7   \\
            A-BIER~\cite{opitz2018deep} & 74.2 & 86.9 & 94.0 & 97.8  \\
            ABE-8~\cite{kim2018attention} & 76.3 & 88.4 & 94.8 & 98.2   \\
            \hline
            BDB & {\bf 83.0 } &  {\bf 93.3 } & {\bf 97.3 }& {\bf 99.2 } \\
            \hline
       \end{tabular}
       }
        \caption{Stanford online products dataset}
    \end{subtable}
    \vspace{-2mm}
    \caption{The comparison on Recall@$K$(\%) scores with other state-of-the-art metric learning methods on CUB200-2011 (cropped), CARS196 (cropped), In-Shop Clothes Retrieval, and Stanford online products datasets.}
    \label{tab:compare_different_dataset}
    \vspace{-3mm}
\end{table*}

\begin{table}
    \begin{center}
        \resizebox{0.9\linewidth}{!}
        {
            \begin{tabular}{c|c|c|c|c}
             \hline
             $K$ & 1 & 5 & 10 & 20 \\
             \hline\hline
             Baseline + LiftedStruct~\cite{oh2016deep} & 66.8 & 88.5 & 93.4 & 96.3 \\
             BDB + LiftedStruct~\cite{oh2016deep} & 71.4 & 89.7 & 93.9 & 96.3  \\
             \hline\hline
             Baseline + Margin~\cite{wu2017sampling} & 65.7 & 88.1 & 93.1 & 96.4   \\
             BDB + Margin~\cite{wu2017sampling} & 72.0 & 90.8 & 94.4 & 97.0   \\
             \hline\hline
             Baseline + Histogram~\cite{ustinova2016histogram} & 64.6 & 87.2 & 93.0 & 96.4   \\
             BDB + Histogram~\cite{ustinova2016histogram} & 73.1 & 90.7 & 94.2 & 96.9   \\
             \hline\hline
             Baseline + Hard Triplet~\cite{hermans2017defense} & 69.5 & 89.5 & 94.0 & 96.8 \\
             BDB + Hard Triplet~\cite{hermans2017defense} & {\bf 74.1} & {\bf 91.0} & {\bf 94.7} & {\bf 97.1} \\
             \hline
            \end{tabular}
        }
    \end{center}\vspace{-4mm}
    \caption{The BDB network performance on the other standard loss functions of metric learning methods. The statistics are based on the CUB200-2011 (cropped) dataset. ``Baseline'' refers to the ResNet-50 Baseline defined in section~\ref{sec:BFE}.}
    \vspace{2mm}
    \label{tab:compare_loss}
\end{table}

\begin{figure}
\includegraphics[width=1.0\linewidth]{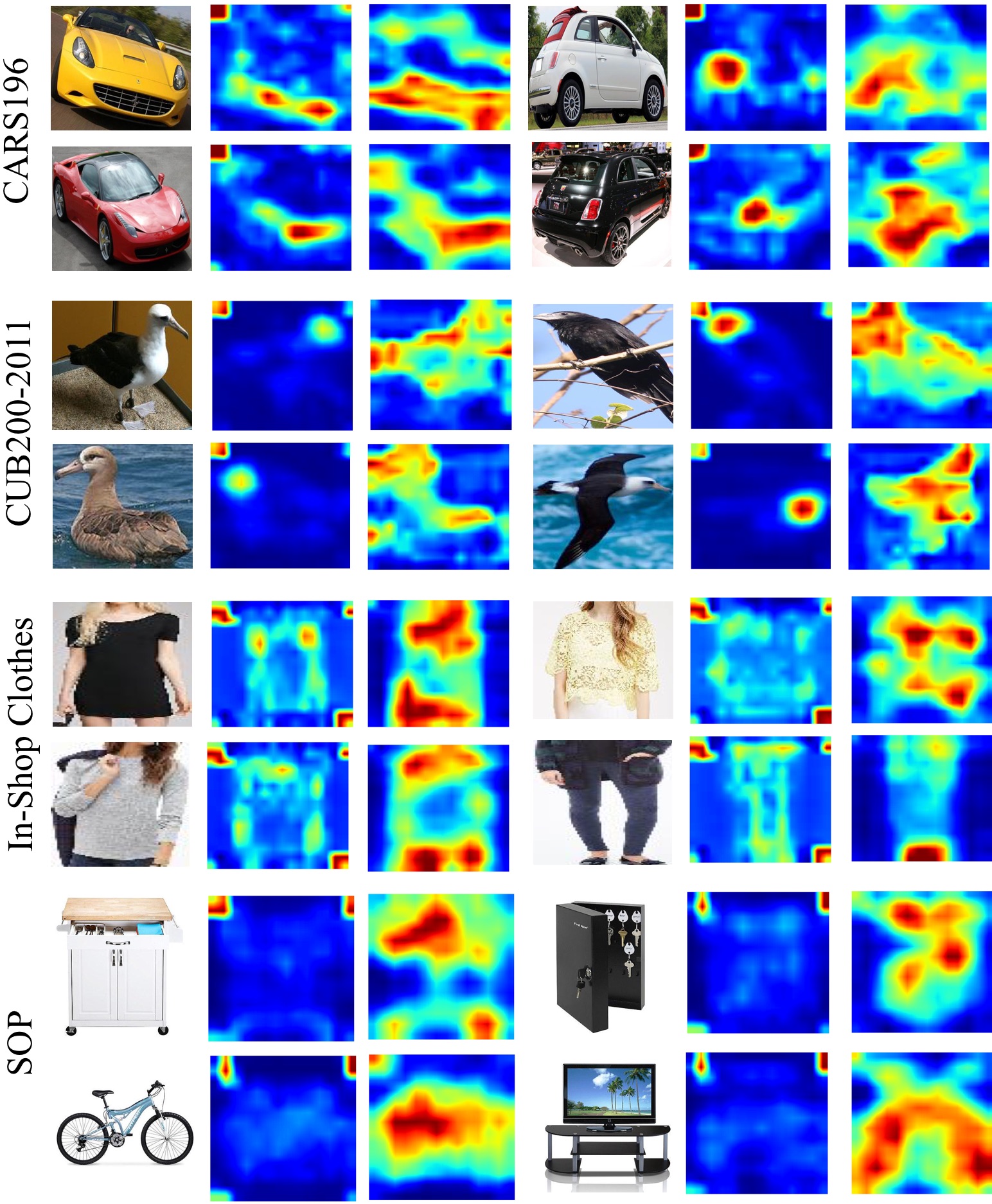}
\caption{The class activation map of Baseline and BDB Network on CARS196, CUB200-2011, In-Shop Clothes retrieval and SOP datasets.}
\label{fig:cub_attention}
\vspace{3mm}
\end{figure}

\begin{figure}
\vspace{-3mm}
\includegraphics[width=1.0\linewidth]{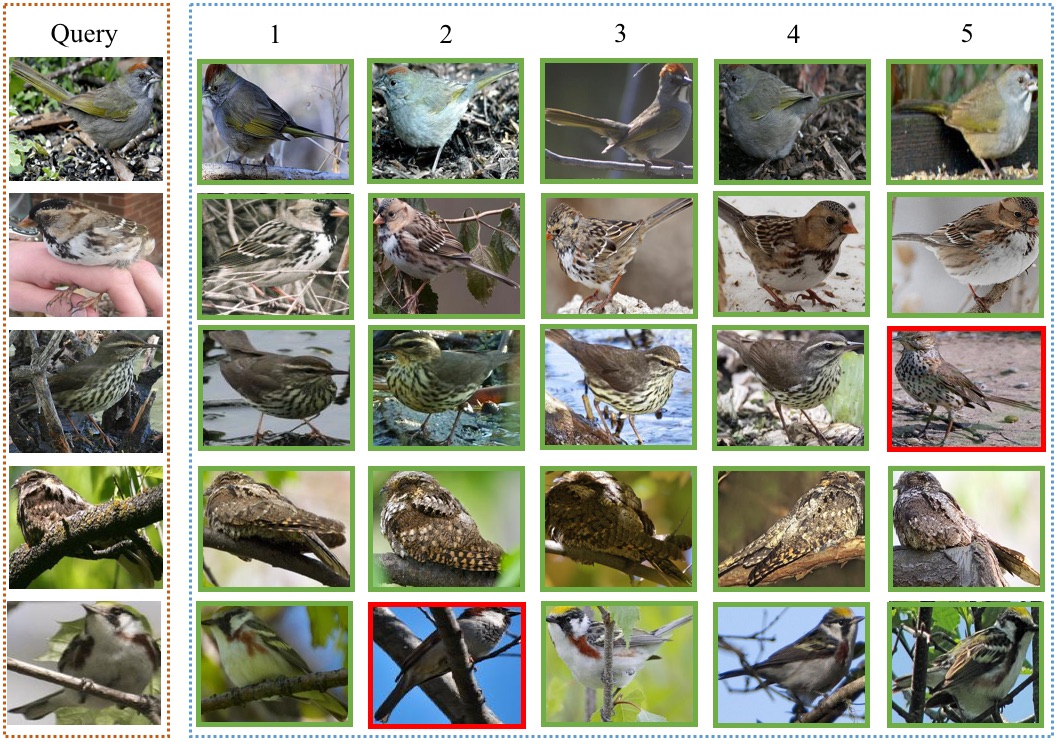}
\vspace{-3mm}
\caption{The top-5 ranking list for the query images on CUB200-2011 dataset from BDB Network. The green and red borders respectively denote the correct and incorrect results.}
\label{fig:cub_retrieval_result}
\vspace{2mm}
\end{figure}

\subsection{Image Retrieval Experiments}
The BDB Network structure can be applied directly on image retrieval problems.
\vspace{-3mm}
\subsubsection{Datasets and Settings}
Our method is evaluated on the commonly used image retrieval datasets including CUB200-2011~\cite{wah2011caltech}, CARS196~\cite{krause20133d}, Stanford online products (SOP)~\cite{oh2016deep}, and In-Shop Clothes retrieval~\cite{liu2016deepfashion} datasets. 
For CUB200-2011 and CARS196, the cropped datasets are used since our BDB Network requires input images to be roughly aligned. 
The experimental setup is the same as that in~\cite{oh2016deep}.
We show the statistics of the four image retrieval datasets in Table~\ref{tab:retrieval_dataset}.


The training images are padded and resized to 256 $\times$ 256 while the aspect ratio is fixed, and then cropped to 224 $\times$ 224 randomly. During testing, CUB200-2011, In-Shop Clothes retrieval dataset, and SOP images are padded on the shorter side and then scaled to 256 $\times$ 256, while CARS196 images are scaled to 256 $\times$ 256 directly. The dropping height ratio and width ratio are both set to 0.5 in the Batch DropBlock Layer. We use the standard Recall@$K$ metric to measure the image retrieval performance. 

\vspace{-3mm}
\subsubsection{Comparison with State-of-the-Art}\vspace{-1mm}
Table~\ref{tab:compare_different_dataset} shows that our BDB Network achieves the best Recall@$1$ scores on all the experimental image retrieval datasets. 
In particular, the BDB Network achieves an obvious improvement (+3.5\%) on the small scale CUB200-2011 dataset which is also the most challenging one. 
On the large scale Stanford online products dataset which contains $22,634$ classes with $120,053$ product images, our BDB network surpasses the state-of-the-art by 6.7\%.
We can see that our BDB Network is applicable on both small and large scale datasets. 

Figure~\ref{fig:cub_retrieval_result} visualizes sample retrieval results of CUB200-2011 (cropped) dataset.
In Figure~\ref{fig:attention}, we also present the class activation maps of Baseline and our BDB network on the CARS196 and CUB200-2011 data-sets.
We can see that our two-branch network encodes more comprehensive features with attentive detail features.
This helps to explain why our BDB Network is in some terms robust to the variance in illumination, poses and occlusions.
\vspace{-3mm}
\subsubsection{Adapt to Other Metric Learning Methods}\vspace{-1mm}
Table~\ref{tab:compare_loss} shows that our BDB Network can also be used with other standard metric learning loss functions, such as lifted structure loss\cite{oh2016deep}, weighted sampling margin loss\cite{wu2017sampling}, and histogram loss\cite{ustinova2016histogram} to boost their performance. 
For a fair comparison, we re-implement the above loss functions on our ResNet-50 Baseline and BDB Network to evaluate their performances. 
Here, the only difference between ResNet-50 Baseline and BDB Network is that the BDB Network has an additional feature dropping branch.
For weighted sampling margin loss, although the ResNet-50 Baseline outperforms the results reported in the work~\cite{wu2017sampling} (+1.8\%), the BDB Network can still improve the result by a large margin (+7.7\%).
We can therefore conclude that the proposed BDB Network can be easily generalized to other standard loss functions in metric learning. 

\section{Conclusion}
In this paper, we propose the Batch DropBlock to improve the optimization in training a neural network for person re-ID and other general metric learning tasks. 
The corresponding BDB Network, which adopts this proposed training mechanism, leverages a global branch to embed salient representations and a feature erasing branch to learn detailed features.
Extensive experiments on both person re-ID datasets and image retrieval datasets show that the BDB Network can make significant improvement on person re-ID and other general image retrieval benchmarks.

{\small
\bibliographystyle{ieee_fullname}
\bibliography{egbib}
}

\end{document}


\title{Batch DropBlock Network for Person Re-identification and Beyond \\(Supplementary Material)}
\maketitle

\section{More Experimental Results}

\paragraph{Dataset without rough alignment}
The prerequisite of Batch DropBlock Network is that the input images are roughly aligned.
To verify it, we evaluate BDB Network on both CUB200 and CARS196 datasets without cropping in which the datasets are not roughly aligned.
We show the results in Table~\ref{tab:uncropped}.
We can see that BDB Network outperforms ResNet-50 Baseline Network due to the two-branch architecture, which corresponds to a feature embedding with a higher dimension.
However, BDB Network without Batch DropBlock ($r_h=0, r_w=0$) performs better than the results with Batch DropBlock ($r_h=0.5, r_w=0.5$).

\begin{table}[h]
    \begin{center}
    \begin{tabular}{c|c|c}
        \hline
         Method & CUB200 & CARS196  \\
         \hline
         ResNet-50 Baseline & 60.1 & 82.3 \\
         $r_h=0.5, r_w=0.5$ & 64.6 & 86.6  \\
         $r_h=0, r_w=0$ & 67.8 & 87.8 \\
         \hline
    \end{tabular}
    \end{center}
    \caption{The Recall@1 score on original CUB200-2011 and CARS196 datasets without cropping. $r_h$ is the erasing height ratio. $r_w$ is the erasing width ratio.}
    \label{tab:uncropped}
\end{table}

\paragraph{Re-ranking}
In Table~\ref{tab:re-ranking}, we compare the results of BDB Network with and without re-ranking~\cite{zhong2017re}. 
It shows that the re-ranking policy can be adopted jointly with our proposed BDB network and it can further improve the final result measured by Rank-1 score and mAP.

\begin{table}[h]
    \begin{center}
        \resizebox{1.0\linewidth}{!}{
    \begin{tabular}{c|c|c|c|c|c|c}
        \hline
         & \multicolumn{2}{c|}{BDB} & \multicolumn{2}{|c}{BDB (re-rank)}& \multicolumn{2}{|c}{BDB+Cut (re-rank)} \\
         Dataset & Rank-1 & mAP & Rank-1 & mAP & Rank-1 & mAP \\
         \hline
         CUHK03-Label  & 73.6 & 71.7 & 83.5 & 85.0 & 87.4 & 88.7 \\
         CUHK03-Detect & 72.8 & 69.3 & 82.0 & 82.9 & 84.7 & 86.5\\
         DukeMTMC-reID & 86.8 & 72.1 & 89.8 & 86.0 & 91.7 & 89.1 \\
         Market1501 & 94.2 & 84.3 & 95.2 & 86.8 & 95.8 & 93.7\\
         \hline
    \end{tabular}
    }
    \end{center}
    \caption{The re-ranked results of BDB Network on person re-ID datasets. }
    \label{tab:re-ranking}
\end{table}

\section{Sample Results}
Figure~\ref{fig:cuhk_results_1} and Figure~\ref{fig:cuhk-cam-sup} shows some challenging queries and more class activation maps from CUHK03 dataset.

More class activation maps from image retrieval datasets are visualized in Figure~\ref{fig:cub-cam-sup}~\ref{fig:car-cam-sup}~\ref{fig:clothes-cam-sup}~\ref{fig:product-cam-sup}.

\begin{figure*}
    \centering
    \includegraphics[width=1.0\linewidth]{Figures/cuhk_results_1.png}
    \caption{Some challenging CUHK03 queries. Red boxes indicate the incorrect results. In the first and fourth row, our method can retrieve the correct person even given a back view query image, which shows our method learns pose invariant feature. The second and third examples show our method learns feature robust to occlusion.}
    \label{fig:cuhk_results_1}
\end{figure*}

\begin{figure*}
    \centering
    \includegraphics[width=1.0\linewidth]{Figures/cam-cuhk-sup.png}
    \includegraphics[width=1.0\linewidth]{Figures/sup-sub-2.png}

    \caption{The comparison of class activation maps between Baseline and the proposed BDB Network on CUHK03 dataset. As shown above, our method learns more spatially distributed and attentive feature representations.}
    \label{fig:cuhk-cam-sup}
\end{figure*}

\begin{figure*}
    \centering
    \includegraphics[width=1.0\linewidth]{Figures/cam-cub-sup.png}
        \includegraphics[width=1.0\linewidth]{Figures/sup-sub.png}
    \caption{The comparison of class activation maps between Baseline and the proposed BDB Network on CUB200 class dataset. We can see that the baseline mainly focuses on head while our method also learns attentive feature from body part.}
    \label{fig:cub-cam-sup}
\end{figure*}

\begin{figure*}
    \centering
    \includegraphics[width=1.0\linewidth]{Figures/cam-car-sup.png}
    \includegraphics[width=1.0\linewidth]{Figures/sup-sub.png}
    \caption{The comparison of class activation maps between Baseline and the proposed BDB Network on CAR196 dataset. We can see that the baseline feature maps contain outliers in background and mainly focus few descriminative regions of the cars. While the features from BDB Network are more spatially ditributed and contain some local saliency.}
    \label{fig:car-cam-sup}
\end{figure*}

\begin{figure*}
    \centering
    \includegraphics[width=1.0\linewidth]{Figures/cam-clothes-sup.png}
        \includegraphics[width=1.0\linewidth]{Figures/sup-sub.png}
    \caption{The comparison of class activation maps between Baseline and the proposed BDB Network on In-Shop Clothes retrieval dataset. We can see that the feature representations from BDB Network have more clear contours.}
    \label{fig:clothes-cam-sup}
\end{figure*}

\begin{figure*}
    \centering
    \includegraphics[width=1.0\linewidth]{Figures/cam-product-sup.png}
        \includegraphics[width=1.0\linewidth]{Figures/sup-sub.png}
    \caption{The comparison of class activation maps between Baseline and the proposed BDB Network on Stanford Online Shopping Product dataset. We can see that the baseline feature maps contain outliers in background while the proposed BDB Network learns features mainly focusing on products.}
    \label{fig:product-cam-sup}
\end{figure*}

{\small
\bibliographystyle{ieee}
\bibliography{egbib}
}